\def\year{2022}\relax
\documentclass[letterpaper]{article} 
\usepackage{aaai22}  
\usepackage{times}  
\usepackage{helvet}  
\usepackage{courier}  
\usepackage[hyphens]{url}  
\usepackage{graphicx} 
\usepackage{natbib}  
\usepackage{caption} 
\DeclareCaptionStyle{ruled}{labelfont=normalfont,labelsep=colon,strut=off} 
\usepackage{algorithm}
\usepackage{algorithmic}

\usepackage{newfloat}
\usepackage{listings}
\floatstyle{ruled}
\newfloat{listing}{tb}{lst}{}
\floatname{listing}{Listing}
\nocopyright
\usepackage{bbding}
\usepackage{pifont}
\usepackage{amssymb}
\usepackage{amsmath}
\usepackage{multirow}
\usepackage{subfig}
\graphicspath{{figures/}}

\title{CMTR: Cross-modality Transformer for Visible-infrared Person Re-identification}
\author{
    Tengfei Liang\textsuperscript{\rm 1},
    Yi Jin\textsuperscript{\rm 1}$^{\dag}$, 
    Yajun Gao\textsuperscript{\rm 1},
    Wu Liu\textsuperscript{\rm 2},
    Songhe Feng\textsuperscript{\rm 1},
    Tao Wang\textsuperscript{\rm 1},
    Yidong Li\textsuperscript{\rm 1}
}
\affiliations{

    \textsuperscript{\rm 1}School of Computer and Information Technology, Beijing Jiaotong University, Beijing, China\\
    \textsuperscript{\rm 2}JD AI Research, Beijing, China\\
    
    \{tengfei.liang, yjin, yajun.gao\}@bjtu.edu.cn, liuwu1@jd.com, \{shfeng, twang, ydli\}@bjtu.edu.cn

}

\begin{document}

\maketitle

\newcommand\blfootnote[1]{%
\begingroup
\renewcommand\thefootnote{}\footnote{#1}%
\addtocounter{footnote}{-1}%
\endgroup
}
\blfootnote{$^{\dag}$Corresponding author.}

\begin{abstract}
Visible-infrared cross-modality person re-identification is a challenging ReID task, which aims to retrieve and match the same identity's images between the heterogeneous visible and infrared modalities. 
Thus, the core of this task is to bridge the huge gap between these two modalities.
The existing convolutional neural network-based methods mainly face the problem of insufficient perception of modalities' information, and can not learn good discriminative modality-invariant embeddings for identities, which limits their performance. 
To solve these problems, we propose a cross-modality transformer-based method (CMTR) for the visible-infrared person re-identification task, which can explicitly mine the information of each modality and generate better discriminative features based on it. 
Specifically, to capture modalities’ characteristics, we design the novel modality embeddings, which are fused with token embeddings to encode modalities’ information.
Furthermore, to enhance representation of modality embeddings and adjust matching embeddings’ distribution, we propose a modality-aware enhancement loss based on the learned modalities’ information, reducing intra-class distance and enlarging inter-class distance.  
To our knowledge, this is the first work of applying transformer network to the cross-modality re-identification task. 
We implement extensive experiments on the public SYSU-MM01 and RegDB datasets, and our proposed CMTR model’s performance significantly surpasses existing outstanding CNN-based methods. 
\end{abstract}

\section{Introduction}

\begin{figure}[t]
\centering
\includegraphics[width=\linewidth]{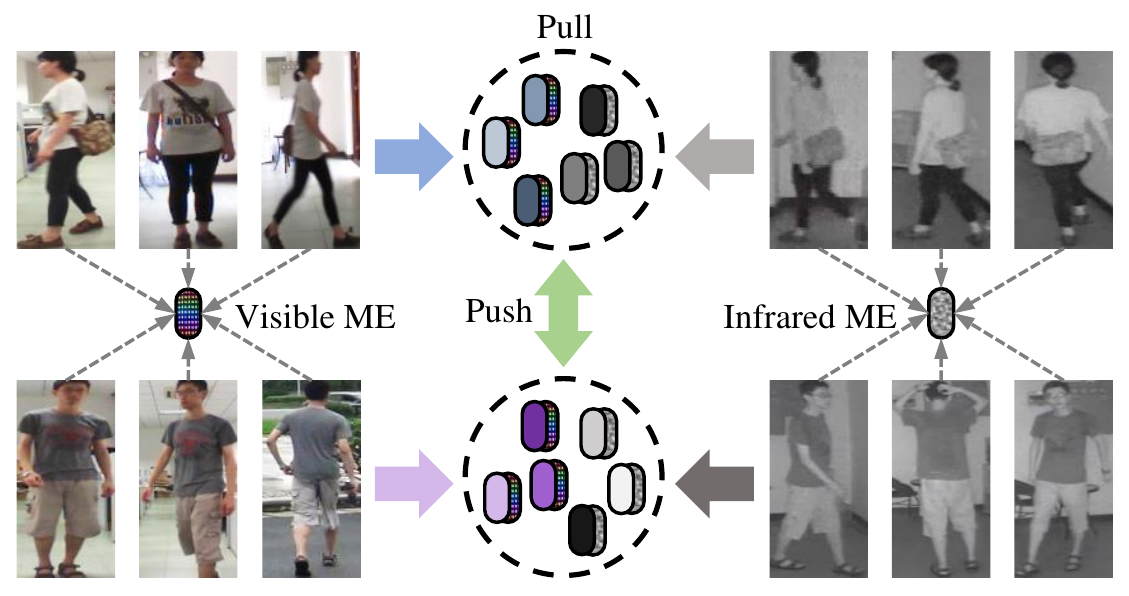}
\caption{Conceptual illustration of CMTR's strategy. It explicitly perceives modality information and generates corresponding embeddings. Based on enhanced modality embeddings (ME), it can pull intra-class features and push inter-class features to learn effective embeddings for retrieval.}
\label{fig_conception}
\end{figure}

Person Re-Identification (ReID) task aims to retrieve the given person's images among multiple different cameras with viewpoint and illumination changes, pose variations, etc \cite{DBLP:journals/corr/ZhengYH16}. 
It has been studied for many years, and the corresponding methods achieve good performance \cite{DBLP:conf/cvpr/0004GLL019, DBLP:conf/mm/WangYCLZ18, DBLP:journals/tpami/9336268}. 
However, existing ReID methods mainly focus on the person retrieval in the single visible modality under RGB cameras, which constrains methods to be used only during the daytime. 
To achieve full-time intelligent video surveillance, based on the mechanism of existing surveillance cameras to automatically switch to infrared mode at night, the visible-infrared cross-modality person ReID (VI-ReID) task \cite{DBLP:conf/iccv/WuZYGL17, DBLP:journals/sensors/NguyenHKP17} is recently proposed to expand the application scope and attracts increasing researchers' attention in this field. 
The VI-ReID requires methods that can match images of the same identity between the visible and infrared modalities, which is more challenging because of the huge heterogeneous gap.

The visible and infrared images are generated by cameras that capture light in different wavelength ranges. 
The former consists of three channels (red, green, and blue) with the color information, while the latter only contains one channel with infrared light radiation. 
They are intrinsically heterogeneous and different. 
To reduce the huge modality gap, a natural strategy proposed by researchers is to transform images of one modality into another. 
There are some GAN-based methods \cite{DBLP:conf/ijcai/DaiJWWH18, DBLP:conf/cvpr/WangWZCS19, DBLP:conf/aaai/WangZYCCLH20, DBLP:conf/iccv/WangZ0LYH19} that attempt to learn the modality translation mapping. 
However, due to the heterogeneous imaging process, the same gray in infrared images can be totally different colors in visible images. 
Therefore, there is no reliable mapping relationship to support the generative model. 
Some recent methods have turned attention to the structural design of the convolutional neural network (CNN) model for more effective feature extraction.
Based on the two-stream architecture, some models (Ye et al. 2018; Hao et al. 2019b; Liu et al.2020; Ye et al. 2021) are designed to use shallow layers with unshared weights to extract shared features for different modalities and use deep layers with shared weights to learn discriminative features. 
However, this strategy cannot well guarantee the envisaged layers’ function and ignore the sufficient perception and deeper mining of the built-in modality characteristics, resulting in limited performance.

\begin{figure}[t]
\centering
\includegraphics[width=\linewidth]{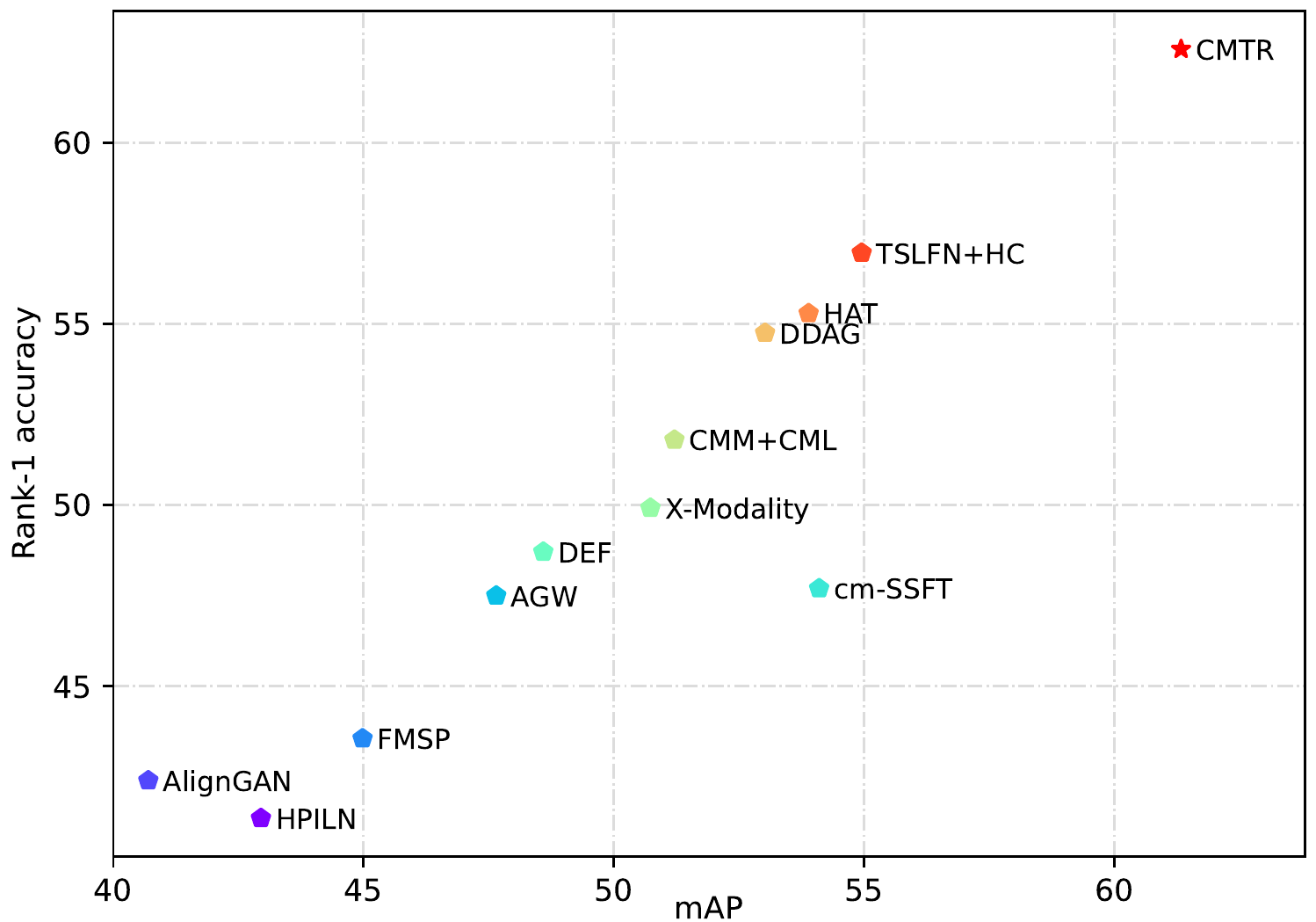}
\caption{Comparison with the existing CNN-based methods on SYSU-MM01 dataset. The Rank-1 and mAP of our CMTR model surpasses the others by a large margin on the most difficult single-shot setting of all-search mode.}
\label{fig_comparison_with_sota}
\end{figure}

To solve the aforementioned problems and break through the limitations of the CNN-based approaches, we explore the new architecture.
Compared with CNN, transformer model shows advantages in single-modality ReID \cite{DBLP:journals/corr/abs-2102-04378}, obtaining global receptive field with self-attention modules and complete spatial features without pooling layers, but it still can't solve the gap problem of this cross-modality task. 
In this paper, we propose the cross-modality transformer (CMTR) model, which can capture modality characteristics by learnable embeddings in an explicit way and generate more effective matching embeddings on this basis. 
Specifically, to mine modalities' characteristics, we first introduce the modality embeddings (ME) to our method. 
Similar to the idea of position embeddings in plain transformer \cite{DBLP:conf/nips/VaswaniSPUJGKP17}, our ME can be integrated into the input phase of the transformer framework by adding to the patches' token embeddings. 
As shown in Figure \ref{fig_conception}, corresponding to the visible and infrared modality, we define two learnable embeddings. 
They are used to learn information of each modality, which can be helpful to the subsequent learning process of modality-invariant embeddings. 
To  enhance the constraint on the learnable ME and optimize matching embeddings' distribution, we further design a novel loss function, the modality-aware enhancement (MAE) loss. 
It consists of the modality-aware center loss and the modality-aware ID loss with modality removal process by subtracting the learned modalities' knowledge from the ME, trying to pull intra-class features and push inter-class features. 

Compared with the state-of-the-art models, our proposed CMTR method shows markedly outstanding performance (as shown in Figure \ref{fig_comparison_with_sota}).
In general, the contributions of our paper can be mainly summarized as follows:

\begin{itemize}
\item We propose a new cross-modality transformer (CMTR) network, which is the first transformer-based exploration for the visible-infrared person re-identification task.
\item We introduce the learnable modality embeddings (ME) to the CMTR network, which directly mine modalities' information and can be used effectively to alleviate the gap between the heterogeneous images.
\item We design a novel modality-aware enhancement (MAE) loss function that enforces the ME to capture more helpful characteristics of each modality and assist the generation of discriminative features.
\item Extensive experiments are conducted on SYSU-MM01 and RegDB benchmarks and demonstrate our method's superior performance against the existing methods.
\end{itemize}

\section{Related Work}

\textbf{Visible-infrared Person Re-identification.}
The Visible-infrared person re-identification attempts to recognize visible and infrared images of a person under cameras of different modalities. 
For the first time, Wu et al. clearly defined the VI-ReID task \cite{DBLP:conf/iccv/WuZYGL17}. 
They contributed the challenging large-scale SYSU-MM01 dataset and proposed the basic Zero-Padding method. 
After that, many researchers put forward some new methods. 
For GAN-based methods, Dai et al. designed the cmGAN \cite{DBLP:conf/ijcai/DaiJWWH18} that uses generative adversarial networks (GAN) to learn discriminative common representations for this cross-modality task. 
Wang et al. proposed the Dual-level Discrepancy Reduction Learning (D$^2$RL) method \cite{DBLP:conf/cvpr/WangWZCS19}, trying to handle the unified multi-spectral representation by image translation. 
And Wang et al. introduced the AlignGAN method \cite{DBLP:conf/iccv/WangZ0LYH19} to jointly exploit pixel alignment and feature alignment and reduce intra-modality variations. 
For the two-stream or multi-stream CNN methods, Ye et al. successively proposed the BDTR framework \cite{DBLP:conf/ijcai/YeWLY18}, the MAC model \cite{DBLP:conf/mm/YeLL19}, the two-stream AGW \cite{DBLP:journals/tpami/9336268}, and the DDAG method \cite{DBLP:conf/eccv/YeSCSL20}, using special structure and loss constraints to make them learn discriminative features implicitly. 
Li et al. \cite{DBLP:conf/aaai/LiWHG20} designed a three-stream structure and introduced an auxiliary X modality to pull different modalities' images.
However, the above methods ignore the direct mining and utilization of modality information, which limits their performance. 
Different from them, our method can effectively overcome this limitation with promising performance.

\noindent\textbf{Research of Transformer.}
The Transformer method was first proposed by Vaswani et al. \cite{DBLP:conf/nips/VaswaniSPUJGKP17} to solve machine translation tasks in the field of natural language processing. 
It and its variants dominate the field for a long time. 
In recent years, researchers successfully applied it to lots of visual tasks, showing its superiority to convolutional neural networks, such as image classification \cite{DBLP:conf/iclr/DosovitskiyB0WZ21}, object detection \cite{DBLP:conf/eccv/CarionMSUKZ20}, semantic segmentation \cite{DBLP:conf/cvpr/zheng2021rethinking}, etc.  
Based on the vision transformer model, our proposed method focuses on the modalities' heterogeneous gap of the cross-modality visible-infrared person re-identification, which is the first exploration in this task and achieves good results.

\begin{figure*}[t]
\centering
\includegraphics[width=\linewidth]{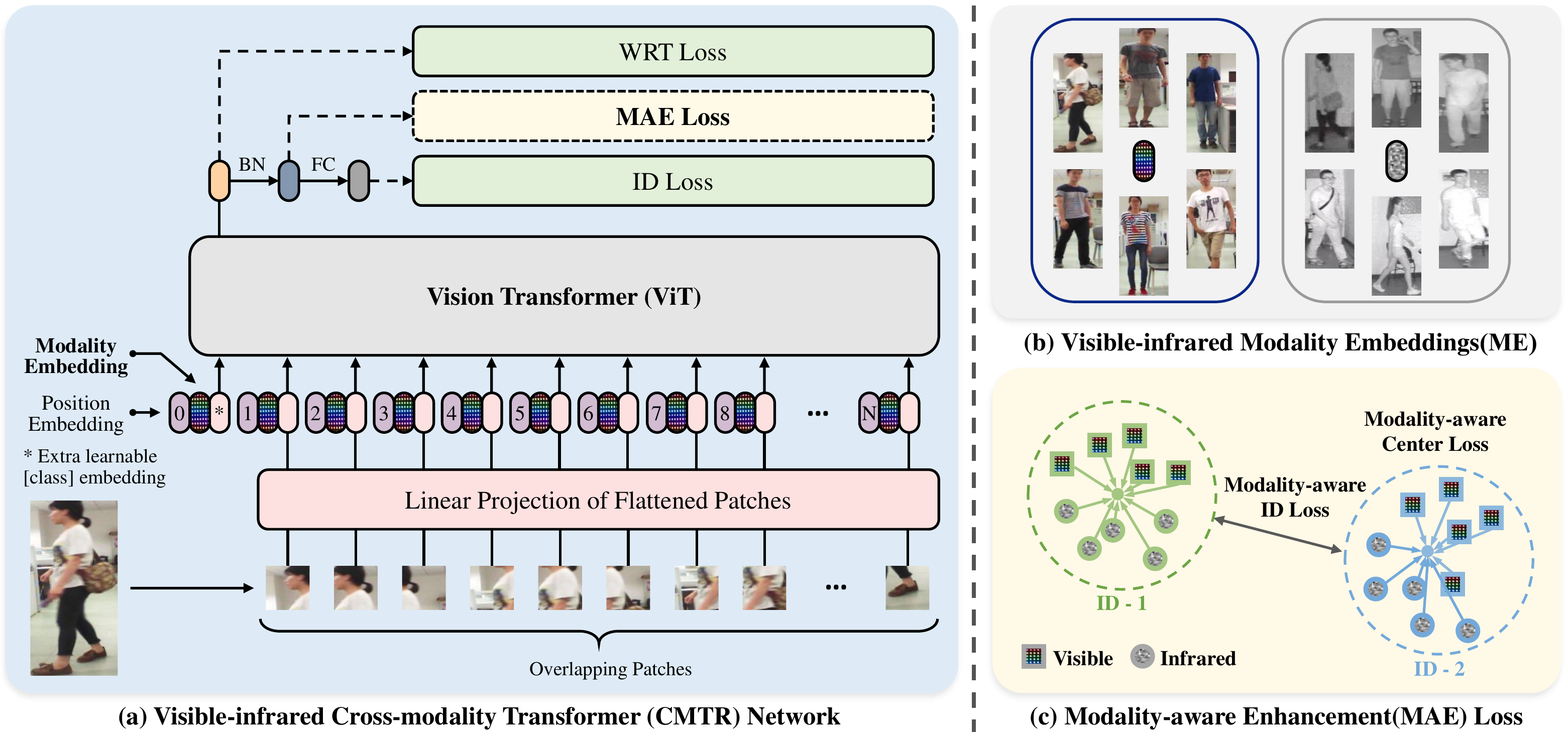}
\caption{The framework of our proposed method. 
(a) The overall structure of visible-infrared cross-modality transformer (CMTR) network, which is based on the Vision Transformer (ViT) backbone with multiple loss constraints. 
(b) The conceptual illustration of the designed visible-infrared modality embeddings (ME). 
(c) The diagram of modality-aware enhancement (MAE) loss, which contains two components, the modality-aware center loss and the modality-aware ID loss. }
\label{fig_overall_structure}
\end{figure*}

\section{The Proposed Method}
In this section, the proposed visible-infrared cross-modality transformer (CMTR) network is explained in detail. 
We introduce the overall network structure in the first subsection. 
Then we focus on the designed modality embeddings and modality-aware enhancement loss and explain their definition and function in the next two subsections. 
In the last subsection of this part, we give the overall formula of the objective function during the process of optimization.

\subsection{Overall Network Structure}
Our CMTR network is built with the vision transformer framework \cite{DBLP:conf/iclr/DosovitskiyB0WZ21}, and we adapt it to the VI-ReID task. 
For input images, we let $vis$ and $ir$ represent the visible modality and infrared modality. 
Thus, the visible image set is denoted as $X^{vis}=\{x^{vis}|x^{vis} \in \mathbb{R}^{C \times H \times W}\}$, and the infrared image set is denoted as $X^{ir}=\{x^{ir}|x^{ir} \in \mathbb{R}^{C \times H \times W}\}$. 
The $C$, $H$, $W$ denote images' channel, height and width respectively. 
In a training batch, there are $B$ images with the same number of $x^{vis}_i$ and $x^{ir}_i$, where $i \in \{1,2,...,B/2\}$. 
As shown in Figure \ref{fig_overall_structure}a, our method mainly contains three stages from the bottom to the top: input embedding, feature extraction, and multi-loss constraint.

In the stage of input embedding, as illustrated at the bottom of Figure \ref{fig_overall_structure}a, here is an example when the input is a visible image (it is similar when inputting an infrared image).
The input image $x^{vis}_i$ is first split into a sequence of patches with the shape of $\mathbb{R}^{N \times C \times P \times P}$, where $P$ denotes the size and $N$ denotes the length of this sequence.
Besides, following \cite{DBLP:journals/corr/abs-2101-11986, DBLP:journals/corr/abs-2102-04378}'s strategy, we generate the patches with overlapping by stride $S$ ($S<P$) (soft  split) to enhance the correlation among adjacent patches. 
The patches are reshaped to flattened embeddings with shape of $\mathbb{R}^{N \times (C \times P^2)}$. 
Then through the linear projection, they are converted to a sequence of token embeddings ($\mathbb{R}^{N \times D}$, $D$ denotes the embedding dimension).
An extra learnable [class] token embedding is merged into the sequence to capture the global attention of the whole image.
In the CMTR network, before being sent to the transformer, the token embedding sequence is fused with position embeddings and the designed modality embeddings (ME). 

In feature extraction stage, the vision transformer (ViT) \cite{DBLP:conf/iclr/DosovitskiyB0WZ21} model is used as the backbone extractor. 
By using multi-layer self-attention modules, the model can perceive more effective global features than CNN-based methods.
As the top of Figure \ref{fig_overall_structure}a shows, corresponding to the location of [class] token, we can get the image vector for each $x^{m}_i$ in the backbone's output.
Let $\mathcal{I}$ and $\mathcal{F}$ denote the process of input embedding and feature extraction. 
The image vector can be extracted as follows:
\begin{equation}
v^{m}_i = \mathcal{F}(\mathcal{I}(x^{m}_i))
\qquad
m \in \{vis, ir\}
\end{equation}

During the last stage of multi-loss constraint, these image vectors $v^{m}_i$ obtained from the training batch pass through batch normalization (BN) layer and fully connected (FC) layer. 
And after these different layers, the method calculates multiple losses, the identity (ID) loss, weighted regularization triplet (WRT) loss and our proposed modality-aware enhancement Loss (MAE) loss, jointly constraining the vectors' distribution to generate more discriminative ID embeddings that are invariant to the visible-infrared modalities.

\subsection{Visible-infrared Modality Embeddings}

\begin{figure}[t]
\centering
\includegraphics[width=\linewidth]{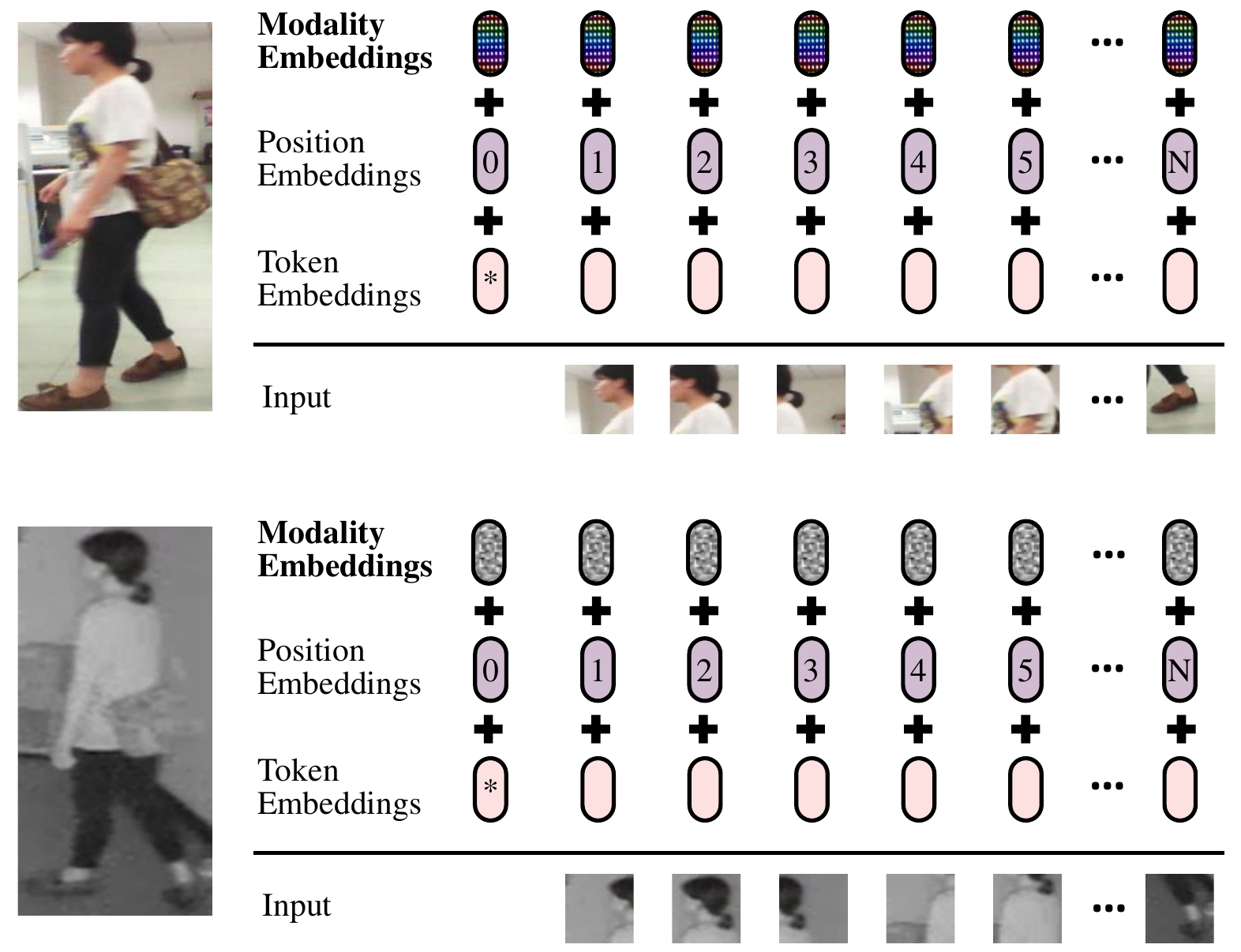}
\caption{The CMTR's input embeddings. The input embeddings are the sum of the token embeddings, the position embeddings and the designed modality embeddings.}
\label{fig_input_embeddings}
\end{figure}

Perception of modality characteristics is helpful to generate modality-invariant features.
However, this key is ignored by many existing methods.
To achieve this, we introduce the modality embeddings (ME) into our CMTR model, which directly aim to learn and capture each modality's inherent information and characteristics.

Inspired by the idea of the position embeddings in transformer \cite{DBLP:conf/nips/VaswaniSPUJGKP17} or the segmentation embeddings in BERT \cite{DBLP:conf/naacl/DevlinCLT19} that can learn positional information or segmented information, our modality embeddings are introduced in a similar way with the different purpose to encode modalities' information. 
This design can be naturally integrated into the transformer framework, which CNN-based models do not have this advantage.
As shown in Figure \ref{fig_input_embeddings}, our CMTR's input embeddings are calculated with three components, i.e., token embeddings, position embeddings and modality embeddings. 
The first two are consistent with previous methods.
For modality embeddings, images in each modality share the same embeddings with all patches.
Specifically, let $\{x^{m}_{i,p1}, x^{m}_{i,p2}, x^{m}_{i,p3}, \cdots, x^{m}_{i,pN}\}$ denote the patch sequence of image $x^{m}_i$. 
The $\{e^{pos}_{p1}, e^{pos}_{p2}, e^{pos}_{p3}, \cdots, e^{pos}_{pN}\}$ denotes position embeddings.
The stage of input embedding $\mathcal{I}(x^{m}_i)$ can be formulated as follows:
\vspace{10pt}
\begin{equation}
\begin{aligned}
\mathcal{I}(x^{m}_i) & = \mathcal{LP}(\{x^{m}_{i,p1}, x^{m}_{i,p2}, x^{m}_{i,p3}, \cdots, x^{m}_{i,pN}\}) \\
& + \{e^{pos}_{p1}, e^{pos}_{p2}, e^{pos}_{p3}, \cdots, e^{pos}_{pN}\} \\
& + 
\begin{cases}
\{e^{vis}, e^{vis}, e^{vis}, \cdots, e^{vis}\},  & \text{if }m\text{ is } vis. \\
\{e^{ir}~, e^{ir}~, e^{ir}~, ~\cdots, e^{ir}\},  & \text{if }m\text{ is } ir.
\end{cases}
\end{aligned}
\label{eq_input_embedding}
\end{equation}
\vspace{3pt}

\noindent where the $\mathcal{LP}$ denotes the linear projection in Figure \ref{fig_overall_structure}a, converting patches' information into token embeddings.
And the $e^{vis}$ and $e^{ir}$ denote the visible modality embedding and the infrared modality embeddings respectively.

As shown in Equation \ref{eq_input_embedding}, these different types of embeddings are fused together in an additive manner. 
The position embeddings $e^{pos}$ vary among patches, while the modality embeddings $e^{m}$ ($m \in \{vis, ir\}$) vary between images' modalities, perceiving different types of information. 

\subsection{Modality-aware Enhancement Loss}
The above-mentioned approach of how to use the modality embeddings makes them capture modalities' characteristics semantically, but the constraint of this way is relatively weak. 
To further enhance ME's ability to learn the modality information, and let the learned ME assist in generating more effective modality-invariant embeddings, we propose the Modality-aware Enhancement (MAE) loss.

As shown at the top of Figure \ref{fig_overall_structure}, the MAE loss acts on the extracted features after batch normalization (BN), which are used as the matching features during testing.
We let $f^{m}_i = BN(v^{m}_i)$ denote the extracted features. 
The MAE loss consists of two parts (Figure \ref{fig_overall_structure}c): the modality-aware center loss and the modality-aware ID loss, which are designed to pull intra-class features and push inter-class features based on the modality embeddings.
And our MAE loss $\mathcal{L}_{MAE}$ is calculated by adding them as follows:
\begin{equation}
\mathcal{L}_{MAE} = \mathcal{L}_{MAC} + \mathcal{L}_{MAID}
\end{equation}
where $\mathcal{L}_{MAC}$ denotes the modality-aware center loss, and $\mathcal{L}_{MAID}$ denotes the modality-aware ID loss.

For the definition of $\mathcal{L}_{MAC}$, it focuses on reducing the gap between different modalities under the same identity, and utilize the learned knowledge from ME to narrow the intra-class features' distance.
During training, we sample $Q$ identities' images in a batch. 
Each identity contains $K/2$ visible images and $K/2$ infrared images.
Specifically, the $\mathcal{L}_{MAC}$ can be formulated by:
\begin{equation}
\begin{aligned}
\mathcal{L}_{MAC} = \sum^Q_{q=1} \sum^K_{k=1} \log{(1+\exp^{\mathcal{D}(f^{m}_{q,k}-\phi_{m}(e^{m}), f^{m}_{q,c})})} \\
f^{m}_{q,c} = \frac{1}{K} \sum^K_{k=1} (f^{m}_{q,k}-\phi_{m}(e^{m})) \qquad m \in \{vis, ir\}
\end{aligned}
\label{eq_MAC_loss}
\end{equation}
where $f^{m}_{q,k}$ denotes the extracted feature from $q$ identity's $k$ image with $m$ modality.
$\phi_{m}(\cdot)$ denotes the mapping to mine the knowledge of modality embeddings $e^{m}$, and in practice, we use the full connection layer to implement it.
In this formula, we let the $f^{m}_{q,k}$ subtract the corresponding $\phi_{m}(e^{m})$ directly to remove modality-specific information and filter out modality-invariant features.
The $f^{m}_{q,c}$ denotes center feature vector of the $q$ identity, which is the mean value of the image features after modality removal.
The $\mathcal{L}_{MAC}$ pulls the distance between id's image features and its center feature vector, and we use the cosine distance $\mathcal{D}(\cdot,\cdot)$ to measure their difference.
Besides, $\mathcal{L}_{MAC}$ use the soft-margin constraint to avoid setting hyperparameter of the hard margin in traditional distance loss functions \cite{DBLP:conf/cvpr/SchroffKP15, DBLP:conf/cvpr/0004GLL019}.
Through the constraint of modality-aware center loss $\mathcal{L}_{MAC}$, our method extracts more compact cross-modality features for each identity.

The modality-aware ID loss $\mathcal{L}_{MAID}$ aims at learning discriminative features among different identities, which is also based on the learned ME's information, designed to push the distance between ids' image features.
The equation of $\mathcal{L}_{MAID}$ can be formulated as follows:
\begin{equation}
\begin{aligned}
\mathcal{L}_{MAID} = \sum^Q_{q=1} \sum^K_{k=1} CrossEntropy(p^{m}_{q,k},t^{m}_{q,k}) \\
p^{m}_{q,k} = Softmax(FC_{id}(f^{m}_{q,k}-\phi_{m}(e^{m})))
\end{aligned}
\label{eq_MAID_loss}
\end{equation}
where $t^{m}_{q,k}$ denotes the one-hot target label for $q$ identity.
The predicted label $p^{m}_{q,k}$ is calculated from the image features $f^{m}_{q,k}-\phi_{m}(e^{m})$ with the same modality removal process as $\mathcal{L}_{MAC}$.
We use the auxiliary FC layer $FC_{id}$ to generate logits for classification, and the $p^{m}_{q,k}$ is obtained through the $Softmax$ operation on logits.
The $\mathcal{L}_{MAID}$ calculates $CrossEntropy(\cdot,\cdot)$ between predictions and targets, attempting to classify different identities’ input images.
With the modality-aware ID loss $\mathcal{L}_{MAID}$'s constraint, the features extracted by the model are given stronger distinguishing ability to achieve more accurate matching.

By optimizing the modality-aware enhancement loss $\mathcal{L}_{MAE}$, firstly, the network can utilize the modality removal process to enforce ME to mine more useful modality-specific characteristics, which is a more direct way to enhance the ME's representation.
Secondly, the ME-based loss functions can adjust the distribution of feature embeddings to be more discriminative for the image retrieval and less affected by the heterogeneous cross-modality gap.

\subsection{Overall Objective Function}
As shown at the top of Figure \ref{fig_overall_structure}a, our CMTR network is constrained by three kinds of losses, and these constraints are jointly optimized.
The overall objective function can be defined as follows:
\begin{equation}
\mathcal{L}_{overall} = \mathcal{L}_{ID} + \mathcal{L}_{WRT} + \lambda \cdot \mathcal{L}_{MAE}
\end{equation}
where $\mathcal{L}_{ID}$ and $\mathcal{L}_{WRT}$ denote the identity (ID) loss and weighted regularization triplet (WRT) loss respectively, and they are common loss constraints \cite{DBLP:conf/cvpr/0004GLL019, DBLP:journals/tpami/9336268} in ReID task.
Thus, we adopt these losses in similar locations.
As for our proposed modality-aware enhancement loss $\mathcal{L}_{MAE}$, it is added to the formula with weight, and the hyperparameter $\lambda$ can control the proportion of this loss.

\section{Experiments}

\subsection{Datasets and Settings}

\subsubsection{Datasets:}
Our experiments are performed on two public datasets, SYSU-MM01 \cite{DBLP:conf/iccv/WuZYGL17} and RegDB \cite{DBLP:journals/sensors/NguyenHKP17}, which are the standard benchmarks and commonly used by existing methods in VI-ReID task.

SYSU-MM01 is currently the largest and most challenging visible-infrared cross-modality person ReID dataset. 
It totally consists of 29,033 visible images and 15,712 infrared images of 491 identities, which are collected by 4 visible cameras and 2 infrared ones from indoors and outdoors. 
The training set contains 22,258 visible images and 11,909 infrared images of 395 identities, and the testing set contains 96 identitie' images. 
Following \cite{DBLP:conf/iccv/WuZYGL17}, 3,803 infrared images of these testing identities are used to form the query set. 
Corresponding to the single-shot or multi-shot setting, 1 or 10 images of each identity under each visible camera are randomly selected to form the gallery set. 
Besides, there are two testing modes: the all-search mode is evaluated with the indoor and outdoor images, while the indoor-search mode is evaluated with only indoor images.

RegDB is collected by dual aligned visible and far-infrared cameras, including 412 identities' images. 
Each identity has 10 visible images and 10 far-infrared images. 
Consistent with previous methods \cite{DBLP:conf/iccv/WangZ0LYH19, DBLP:conf/cvpr/LuWLZLCY20, DBLP:conf/aaai/Zhao0CLY21}, we equally divide the dataset into two parts as the training set and testing set by random selection. 
Each set contains 2,060 visible images and 2,060 far-infrared images. 
In testing set, when performing Visible to Thermal/Thermal to Visible mode, all the 2,060 visible/far-infrared images are used as query set, and all the 2,060 far-infrared/visible images are used as gallery set.

\subsubsection{Evaluation Metrics:}
We evaluate methods with two widely used metrics of this task: the Cumulative Matching Characteristics (CMC) curve and the mean Average Precision (mAP). 
The CMC is denoted as Rank-k (Rk for short) to measure the correct rate in the k-nearest matching results, and we calculate R1, R10, R20 in the experiments. 
Besides, according to \cite{DBLP:journals/tpami/9336268}, the mean inverse negative penalty (mINP) is also used as an auxiliary metric in our ablation study. 
Following \cite{DBLP:conf/iccv/WuZYGL17}, we conduct repeated random selections of gallery on SYSU-MM01 for 10 times to get the more stable average result. 
Similarly, our experiments on RegDB average the results from 10 times repeated random partition of training and testing sets.

\subsubsection{Implementation Details:}
Our proposed method is implemented with the PyTorch \cite{DBLP:conf/nips/PaszkeGMLBCKLGA19} deep learning framework. 
For the transformer backbone, we use the ViT-Base \cite{DBLP:conf/iclr/DosovitskiyB0WZ21} model with pretrained weights on the ImageNet \cite{DBLP:conf/cvpr/DengDSLL009} dataset. 
Before entering the network, the visible and infrared images are resized to $3 \times 256 \times 128$ ($C \times H \times W$). 
We repeat the infrared image’s single channel three times to make it contain three channels. 
The patches are generated with $16 \times 16$ size following \cite{DBLP:conf/iclr/DosovitskiyB0WZ21}, and the stride $S$ is set to 8 for half overlap. 
During training, we adopt the common data augmentation strategies: the random horizontal flip and random erasing \cite{DBLP:conf/aaai/Zhong0KL020}. 
In a mini-batch, we randomly sample 8 identities’ images and each identity has 4 visible images and 4 infrared images. 
We use the AdamW \cite{DBLP:conf/iclr/LoshchilovH19} optimizer with weight decay set to 0.0005. 
The whole model is totally trained for 70 epochs, and the base learning rate is initialized at 0.001 with decay by 0.1 at epoch 15 and 30. 
Besides, we make all pretrained layers’ learning rate to be 0.1 times of the base learning rate. 
The trade-off hyperparameter $\lambda$ in objective function is empirically set to 4. 
During testing, all the query and gallery images are sent into the model to extract feature embeddings with cosine distance to rank retrieval results.

\begin{table*}[h]
    \centering
    \fontsize{8}{12}\selectfont
    \setlength{\tabcolsep}{0.8mm}
    \begin{tabular}{l|c|c c c c|c c c c|c c c c|c c c c}
        \hline
        \hline
        \multirow{3}{*}{~~~~~~~~~~~~~~~~Methods} & \multirow{3}{*}{Venue} & \multicolumn{8}{c}{\textit{All-Search}} & \multicolumn{8}{|c}{\textit{Indoor-Search}} \cr
        \cline{3-18}
        & ~ & \multicolumn{4}{c}{\textit{Single-Shot}} & \multicolumn{4}{|c}{\textit{Multi-Shot}} & \multicolumn{4}{|c}{\textit{Single-Shot}} & \multicolumn{4}{|c}{\textit{Multi-Shot}} \cr
        & ~ & R1 & R10 & R20 & mAP & R1 & R10 & R20 & mAP & R1 & R10 & R20 & mAP & R1 & R10 & R20 & mAP \cr
        \hline
        cmGAN \cite{DBLP:conf/ijcai/DaiJWWH18} & IJCAI 18 & 26.97  & 67.51  & 80.56  & 27.80  & 31.49  & 72.74  & 85.01  & 22.27  & 31.63  & 77.23  & 89.18  & 42.19  & 37.00  & 80.94  & 92.11  & 32.76 \cr
        D$^2$RL \cite{DBLP:conf/cvpr/WangWZCS19} & CVPR 19 & 28.90  & 70.60  & 82.40  & 29.20  & - & - & - & - & - & - & - & - & - & - & - & - \cr
        Hi-CMD \cite{DBLP:conf/cvpr/ChoiLKKK20} & CVPR 20 & 34.94  & 77.58  &  & 35.94  & - & - & - & - & - & - & - & - & - & - & - & - \cr
        JSIA \cite{DBLP:conf/aaai/WangZYCCLH20} & AAAI 20 & 38.10  & 80.70  & 89.90  & 36.90  & 45.10  & 85.70  & 93.80  & 29.50  & 43.80  & 86.20  & 94.20  & 52.90  & 52.70  & 91.10  & 96.40  & 42.70 \cr
        AlignGAN \cite{DBLP:conf/iccv/WangZ0LYH19} & ICCV 19 & 42.40  & 85.00  & 93.70  & 40.70  & 51.50  & 89.40  & 95.70  & 33.90  & 45.90  & 87.60  & 94.40  & 54.30  & 57.10  & 92.70  & 97.40  & 45.30 \cr
        TS-GAN \cite{DBLP:journals/corr/abs-2007-07452} & ArXiv 20 & 49.80 & 87.30 & 93.80 & 47.40 & 56.10 & 90.20 & 96.30 & 38.50 & 50.40 & 90.80 & 96.80 & 63.10 & 59.30 & 91.20 & 97.80 & 50.20 \cr
        \hline
        Zero-Padding \cite{DBLP:conf/iccv/WuZYGL17} & ICCV 17 & 14.80  & 54.12  & 71.33  & 15.95  & 19.13  & 61.40  & 78.41  & 10.89  & 20.58  & 68.38  & 85.79  & 26.92  & 24.43  & 75.86  & 91.32  & 18.64 \cr
        BDTR \cite{DBLP:conf/ijcai/YeWLY18} & IJCAI 18 & 17.01  & 55.43  & 71.96  & 19.66  & - & - & - & - & - & - & - & - & - & - & - & - \cr
        D-HSME \cite{DBLP:conf/aaai/HaoWLG19} & AAAI 19 & 20.68  & 62.74  & 77.95  & 23.12  & - & - & - & - & - & - & - & - & - & - & - & - \cr
        SDL \cite{DBLP:journals/tcsv/KansalSWS20} & TCSVT 20 & 28.12  & 70.23  & 83.67  & 29.01  & - & - & - & - & 32.56  & 80.45  & 90.67  & 39.56  & - & - & - & - \cr
        MAC \cite{DBLP:conf/mm/YeLL19} & MM 19 & 33.26  & 79.04  & 90.09  & 36.22  & - & - & - & - & - & - & - & - & - & - & - & - \cr
        MSR \cite{DBLP:journals/tip/FengLX20} & TIP 19 & 37.35  & 83.40  & 93.34  & 38.11  & 43.86  & 86.94  & 95.68  & 30.48  & 39.64  & 89.29  & 97.66  & 50.88  & 46.56  & 93.57  & 98.80  & 40.08 \cr
        HPILN \cite{DBLP:journals/iet-ipr/ZhaoLXX19} & IET IP 19 & 41.36  & 84.78  & 94.51  & 42.95  & 47.56  & 88.13  & 95.98  & 36.08  & 45.77  & 91.82  & 98.46  & 56.52  & 53.05  & 93.71  & 98.93  & 47.48 \cr
        FMSP \cite{DBLP:journals/ijcv/WuZGL20} & IJCV 20 & 43.56  & - & - & 44.98  & - & - & - & - & 48.62  & - & - & 57.50  & - & - & - & - \cr
        AGW \cite{DBLP:journals/tpami/9336268} & TPAMI 21 & 47.50  & 84.39  & 92.14  & 47.65  & - & - & - & - & 54.17  & 91.14  & 95.98  & 62.97  & - & - & - & - \cr
        cm-SSFT \cite{DBLP:conf/cvpr/LuWLZLCY20} & CVPR 20 & 47.70  & - & - & 54.10  & 57.40  & - & - & 59.10  & - & - & - & - & - & - & - & - \cr
        DEF \cite{DBLP:conf/mm/HaoWGLW19} & MM 19 & 48.71  & 88.86  & 95.27  & 48.59  & 54.63  & 91.62  & 96.83  & 42.14  & 52.25  & 89.86  & 95.85  & 59.68  & 59.62  & 94.45  & 98.07  & 50.60 \cr
        X-Modality \cite{DBLP:conf/aaai/LiWHG20} & AAAI 20 & 49.92  & 89.79  & 95.96  & 50.73  & - & - & - & - & - & - & - & - & - & - & - & - \cr
        CMM+CML \cite{DBLP:conf/mm/LingZLRLS20} & MM 20 & 51.80  & 92.72  & 97.71  & 51.21  & 56.27  & 94.08  & 98.12  & 43.39  & 54.98  & 94.38  & 99.41  & 63.70  & 60.42  & 96.88  & 99.50  & 53.52 \cr
        DDAG \cite{DBLP:conf/eccv/YeSCSL20} & ECCV 20 & 54.75  & 90.39  & 95.81  & 53.02  & - & - & - & - & 61.02  & 94.06  & 98.41  & 67.98  & - & - & - & - \cr
        HAT \cite{DBLP:journals/tifs/YeSS21} & TIFS 21 & 55.29  & 92.14  & 97.36  & 53.89  & - & - & - & - & 62.10  & 95.75  & 99.20  & 69.37  & - & - & - & - \cr
        TSLFN+HC \cite{DBLP:journals/ijon/ZhuYWZHT20} & NeuroC 20 & 56.96  & 91.50  & 96.82  & 54.95  & 62.09  & 93.74  & 97.85  & 48.02  & 59.74  & 92.07  & 96.22  & 64.91  & 69.76  & 95.85  & 98.90  & 57.81 \cr
        \hline
        \textbf{CMTR (Ours)} & - & \textbf{62.58} & \textbf{93.79} & \textbf{98.01} & \textbf{61.33} & \textbf{68.39} & \textbf{95.73} & \textbf{98.82} & \textbf{55.69} & \textbf{67.02} & \textbf{96.86} & \textbf{99.40} & \textbf{73.78} & \textbf{75.40} & \textbf{98.37} & \textbf{99.52} & \textbf{66.84} \cr
        \hline
        \hline
    \end{tabular}
    \caption{Comparison with CNN-based cross-modality ReID methods on the SYSU-MM01 dataset. We roughly divide the existing methods into two categories: GAN-based methods and two/multi-stream CNN methods with different loss functions.}
    \label{tab_comparison_sysumm01_sota}
\end{table*}

\subsection{Comparison with State-of-the-art Methods}
In experiments, we compare our proposed CMTR model with the existing outstanding CNN-based methods in the visible-infrared cross-modality person ReID task. 
As shown in Table \ref{tab_comparison_sysumm01_sota}, we present the quantitative comparison on the SYSU-MM01 dataset. 
Our CMTR method significantly outperforms the other existing methods in all evaluation modes, including all-search/indoor-search with single-shot/multi-shot modes.
In the most challenging single-shot all-search mode, the CMTR achieves 62.58\% Rank-1 and 61.33\% mAP, and it is worth noting that even compared with the strong model TSLFN+HC \cite{DBLP:journals/ijon/ZhuYWZHT20} using a complicated local blocking strategy, our method has a large improvement (+5.62\% Rank-1, +6.38\% mAP) with only global feature extraction. 
From Figure \ref{fig_comparison_with_sota}, we can also intuitively observe the superior performance of the CMTR.

In Table \ref{tab_comparison_regdb_sota}, we show the comparison results on the RegDB dataset. 
Our CMTR method shows great performance, which achieves 80.62\% (+8.79\%) Rank-1 and 74.42\% (+6.86\%) mAP under Visible to Thermal (V to T) mode and 81.06\% (+11.04\%) Rank-1 and 73.75\% (+7.45\%) mAP under Thermal to Visible (T to V) mode, surpassing the latest HAT \cite{DBLP:journals/tifs/YeSS21} method by a large margin. 
All the experimental results demonstrate the effectiveness and robustness of our proposed method.

\begin{table}[h]
    \centering
    \fontsize{8}{12}\selectfont
    \setlength{\tabcolsep}{0.9mm}
    \begin{tabular}{l|c|c c|c c}
        \hline
        \hline
        \multirow{2}{*}{~~~~~~~~~~~~~~~~Methods} & \multirow{2}{*}{Venue}
        & \multicolumn{2}{c}{\textit{V to T}} & \multicolumn{2}{|c}{\textit{T to V}} \cr
        & ~ & R1 & mAP & R1 & mAP \cr
        \hline
        Zero-Padding \cite{DBLP:conf/iccv/WuZYGL17} & ICCV 17 & 17.75  & 18.90  & 16.63  & 17.82 \cr
        SDL \cite{DBLP:journals/tcsv/KansalSWS20} & TCSVT 20 & 26.47  & 23.58  & 25.74  & 22.89 \cr
        BDTR \cite{DBLP:conf/ijcai/YeWLY18} & IJCAI 18 & 33.47  & 31.83  & 32.72  & 31.10 \cr
        MAC \cite{DBLP:conf/mm/YeLL19} & MM 19 & 36.43  & 37.03  & 36.20  & 36.63 \cr
        HSME \cite{DBLP:conf/aaai/HaoWLG19} & AAAI 19 & 41.34  & 38.82  & 40.67  & 37.50 \cr
        D$^2$RL \cite{DBLP:conf/cvpr/WangWZCS19} & CVPR 19 & 43.40  & 44.10  & - & - \cr
        MSR \cite{DBLP:journals/tip/FengLX20} & TIP 19 & 48.43  & 48.67  & - & - \cr
        JSIA \cite{DBLP:conf/aaai/WangZYCCLH20} & AAAI 20 & 48.50  & 49.30  & 48.10  & 48.90 \cr
        D-HSME \cite{DBLP:conf/aaai/HaoWLG19} & AAAI 19 & 50.85  & 47.00  & 50.15  & 46.16 \cr
        AlignGAN \cite{DBLP:conf/iccv/WangZ0LYH19} & ICCV 19 & 57.90  & 53.60  & 56.30  & 53.40 \cr
        FMSP \cite{DBLP:journals/ijcv/WuZGL20} & IJCV 20 & 65.07  & 64.50  & - & - \cr
        CMM+CML \cite{DBLP:conf/mm/LingZLRLS20} & MM 20 & - & - & 59.81  & 60.86 \cr
        X-Modality \cite{DBLP:conf/aaai/LiWHG20} & AAAI 20 & - & - & 62.21  & 60.18 \cr
        cm-SSFT \cite{DBLP:conf/cvpr/LuWLZLCY20} & CVPR 20 & 65.40  & 65.60  & 63.80  & 64.20 \cr
        DDAG \cite{DBLP:conf/eccv/YeSCSL20} & ECCV 20 & 69.34  & 63.46  & 68.06  & 61.80 \cr
        DEF \cite{DBLP:conf/mm/HaoWGLW19} & MM 19 & 70.13  & 69.14  & 67.99  & 66.70 \cr
        Hi-CMD \cite{DBLP:conf/cvpr/ChoiLKKK20} & CVPR 20 & 70.93  & 66.04  & - & - \cr
        HAT \cite{DBLP:journals/tifs/YeSS21} & TIFS 21 & 71.83  & 67.56  & 70.02  & 66.30 \cr
        \hline
        CMTR (Ours) & - & \textbf{80.62} & \textbf{74.42} & \textbf{81.06} & \textbf{73.75} \cr
        \hline
        \hline
    \end{tabular}
    \caption{Comparison with CNN-based on RegDB dataset.}
    \label{tab_comparison_regdb_sota}
    \vspace{-100pt}
\end{table}

\vspace{100pt}~
\vspace{30pt}~


\begin{figure*}[t]
\vspace{-10pt}
\centering
\subfloat[]{\includegraphics[height=0.185\linewidth]{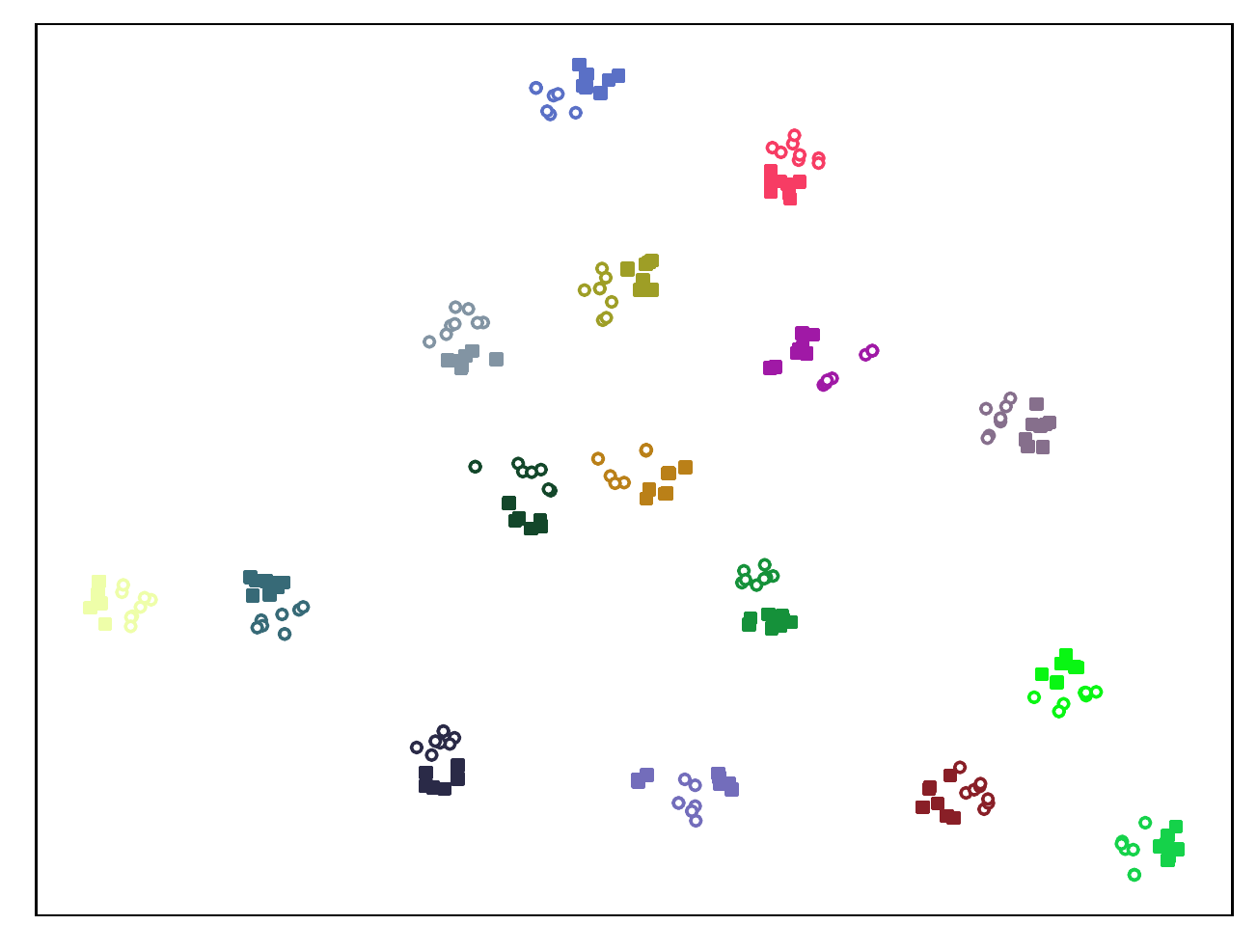} \label{base}}
\subfloat[]{\includegraphics[height=0.185\linewidth]{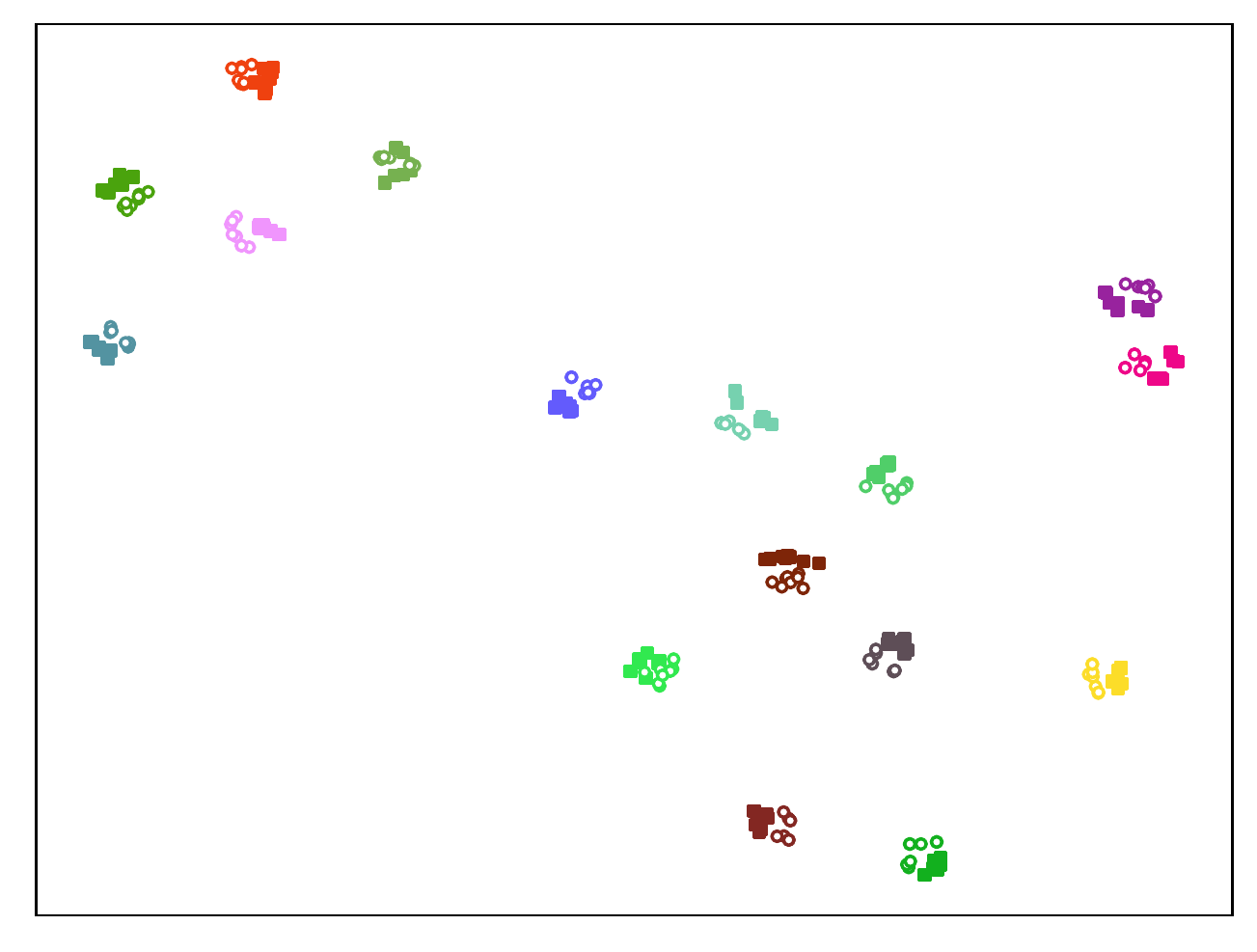} \label{base_me_mac}}
\subfloat[]{\includegraphics[height=0.185\linewidth]{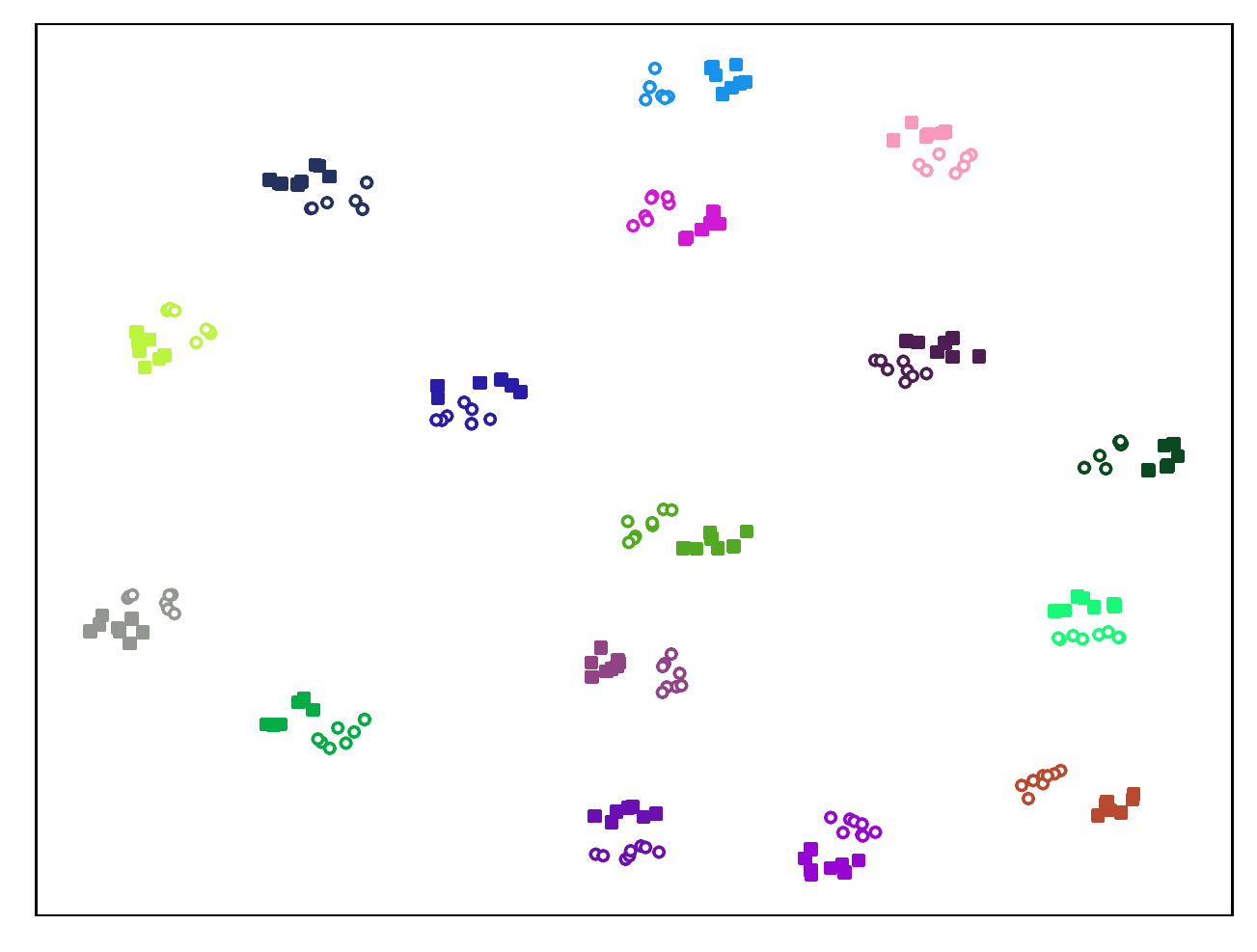} \label{base_me_maid}}
\subfloat[]{\includegraphics[height=0.185\linewidth]{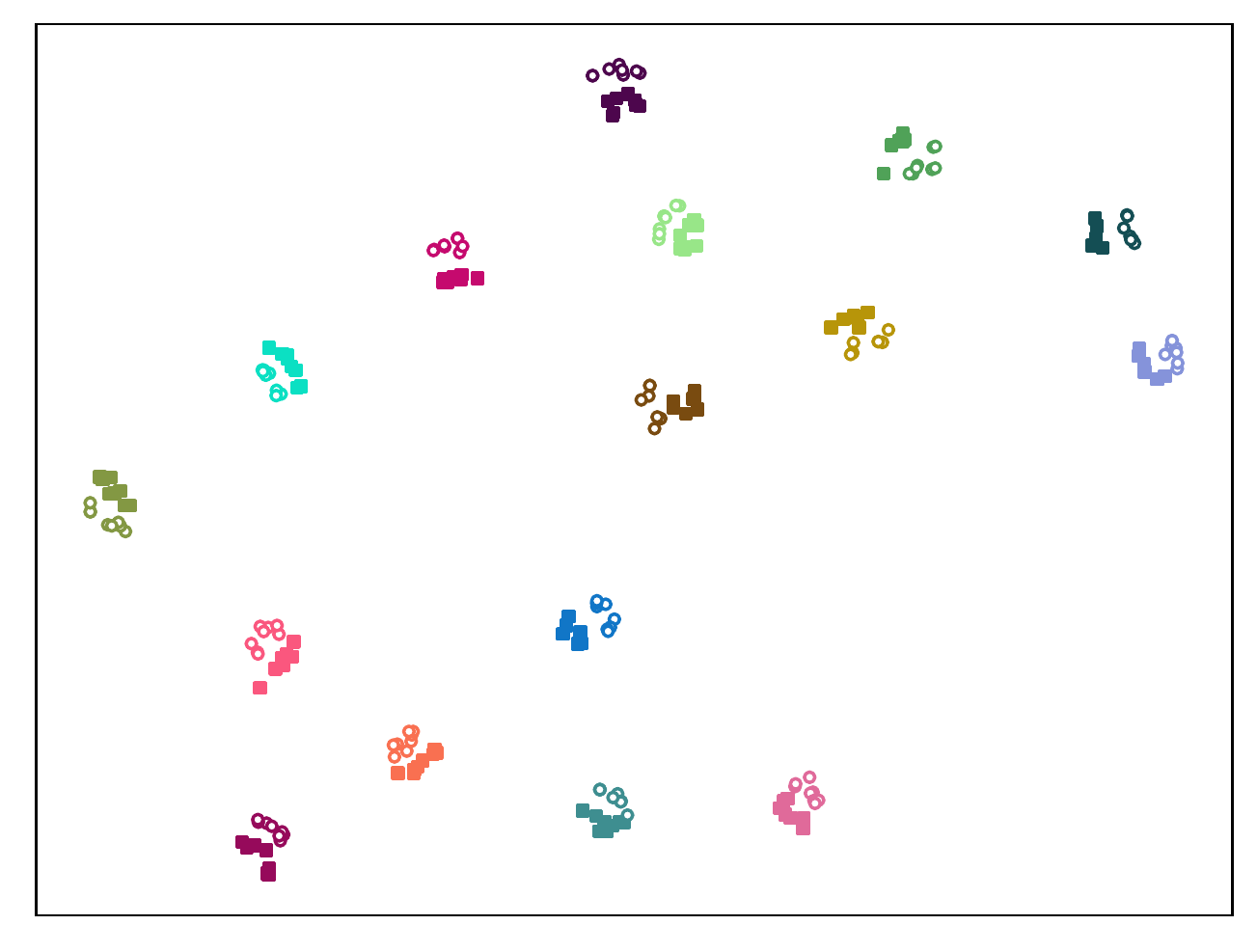} \label{base_me_mac_maid}}
\caption{Feature distributions visualized with t-SNE method. 
(a) Feature distribution of BASE.
(b) Feature distribution of BASE+MAC.
(c) Feature distribution of BASE+MAID.
(d) Feature distribution of BASE+MAE (MAC\&MAID).}
\label{fig_feature_distribution}
\end{figure*}

\begin{table}[h]
    \centering
    \fontsize{8}{12}\selectfont
    \setlength{\tabcolsep}{0.9mm}
    \begin{tabular}{c|c c c c|c c c c c}
        \hline
        \hline
        \multirow{2}{*}{Index} & \multirow{2}{*}{BASE~~} & \multirow{2}{*}{ME}
        & \multicolumn{2}{c|}{~~MAE} & \multirow{2}{*}{R1} & \multirow{2}{*}{R10} & \multirow{2}{*}{R20} & \multirow{2}{*}{mAP} & \multirow{2}{*}{mINP}\cr
        & ~ & ~ & \textit{~~~MAC} & \textit{MAID} \cr
        \hline
        1 & \ding{51} & \ding{55}  & ~\ding{55} & \ding{55} & 54.28 & 90.66 & 96.43 & 53.97 & 41.43 \cr
        2 & \ding{51} & \ding{51}  & ~\ding{55} & \ding{55} & 57.40 & 89.31 & 95.13 & 56.16 & 43.86 \cr
        3 & \ding{51} & \ding{51}  & ~\ding{51} & \ding{55} & 59.53 & 92.63 & 97.22 & 59.40 & 47.90 \cr
        4 & \ding{51} & \ding{51}  & ~\ding{55} & \ding{51} & 60.53 & 91.37 & 96.24 & 58.73 & 45.71 \cr
        \hline
        5 & \ding{51} & \ding{51}  & ~\ding{51} & \ding{51} & \textbf{62.58} & \textbf{93.79} & \textbf{98.01} & \textbf{61.33} & \textbf{49.01} \cr
        \hline
        \hline
    \end{tabular}
    \caption{Ablation studies on the SYSU-MM01 dataset.}
    \label{tab_ablation_base_me_mae}
\end{table}

\subsection{Ablation Study}
The ablation experiments are designed to evaluate the influence of our proposed modality embeddings (ME) and modality-aware enhancement (MAE) loss. 
We conduct these experiments on the SYSU-MM01 dataset with the difficult single-shot setting of all-search mode.

\subsubsection{Effectiveness of the ME and MAE Loss.}
As shown in Table \ref{tab_ablation_base_me_mae}, the "BASE" represents the baseline ViT network trained with the common $\mathcal{L}_{ID}$ and $\mathcal{L}_{WRT}$ (Index-1).
By introducing modality embeddings (ME) into this network, the model (Index-2) achieves 3.12\% Rank-1 and 2.19\% mAP improvements.
For the study of modality-aware enhancement (MAE) loss, we verify the effects of its components $\mathcal{L}_{MAC}$ and $\mathcal{L}_{MAID}$ respectively (Index-3 \& 4).
Compared with "BASE"+"ME", the MAC loss brings +2.13\% Rank-1 and +3.24\% mAP and the MAID loss brings +3.13\% Rank-1 and +2.57\% mAP.
What's more, through jointly optimizing the MAC loss and the MAID loss, i.e., the MAE loss, our method’s performance is further improved (+5.18\% Rank-1 and +5.17\% mAP), which presents the complementarity and potentiality between these two kinds of losses. 

\subsubsection{Analysis of the Modality Embeddings.}
\begin{figure}[t]
\centering
\subfloat[]{\includegraphics[height=0.31\linewidth]{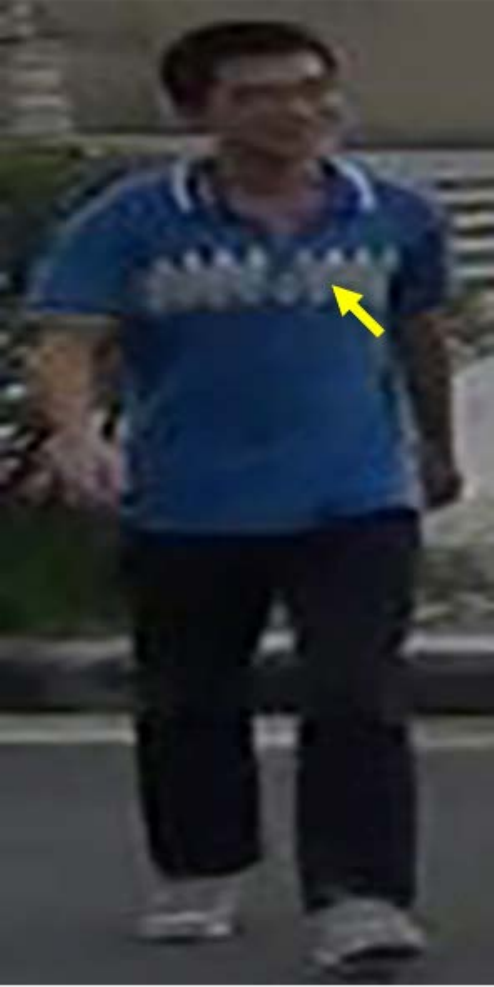} \label{cam_person_vis}}
\subfloat[]{\includegraphics[height=0.31\linewidth]{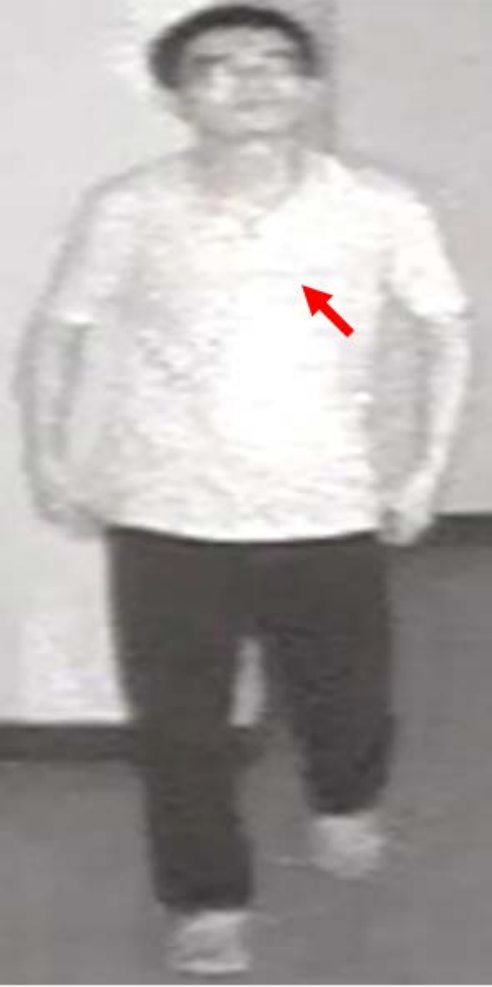} \label{cam_person_ir}}
\subfloat[]{\includegraphics[height=0.31\linewidth]{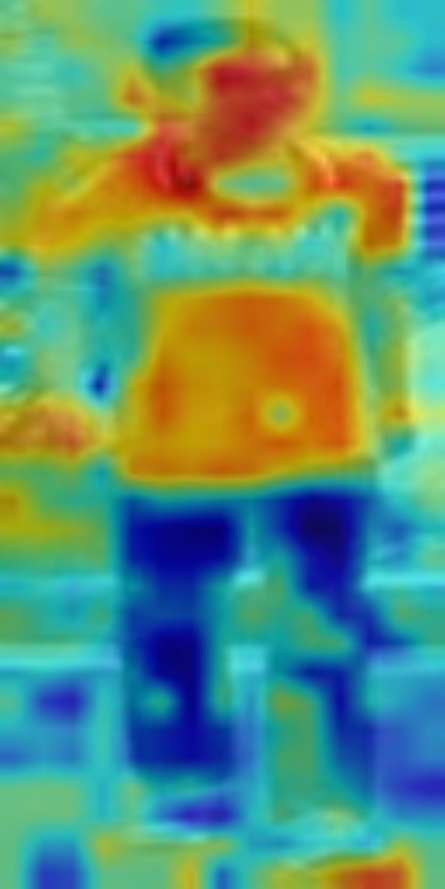} \label{cam_base_id1_vis}}
\subfloat[]{\includegraphics[height=0.31\linewidth]{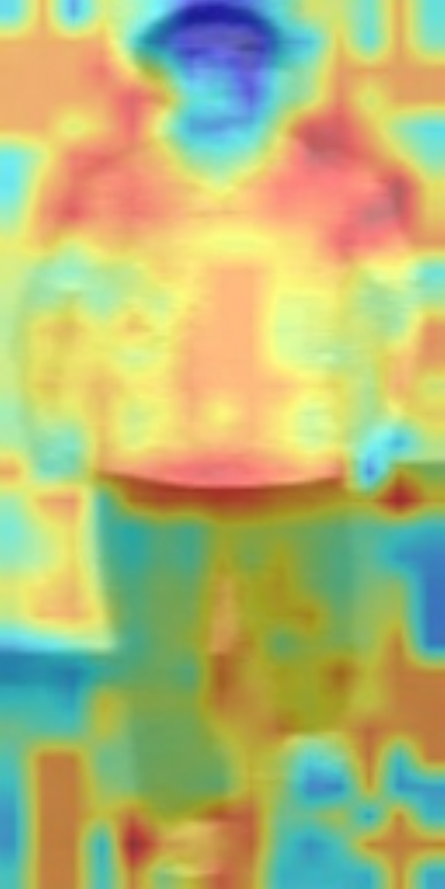} \label{cam_base_id1_ir}}
\subfloat[]{\includegraphics[height=0.31\linewidth]{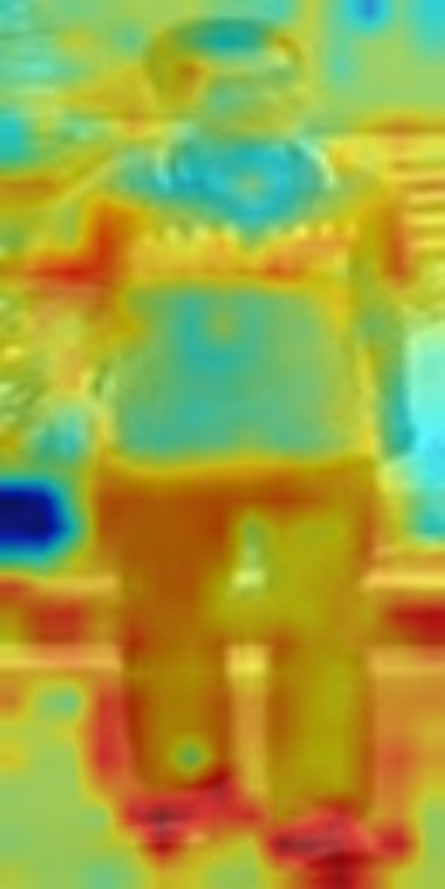} \label{cam_base_me_id1_vis}}
\subfloat[]{\includegraphics[height=0.31\linewidth]{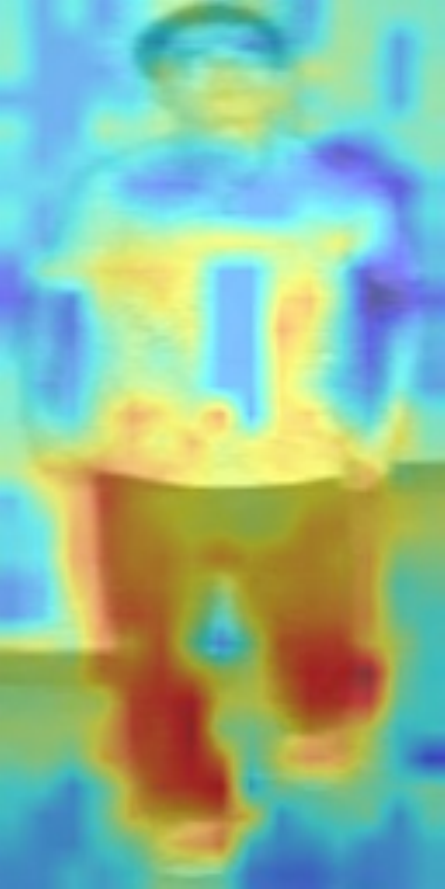} \label{cam_base_me_id1_ir}}
\caption{Visualization with Grad-CAM method. 
(a/b) Visible/Infrared image.
(c/d) BASE's CAM on visible/infrared image.
(e/f) BASE+ME's CAM on visible/infrared image.}
\label{fig_cam_vis_ir}
\end{figure}

To better understand the influence of modality embeddings, we visualized attention maps of the BASE and BASE+ME models. 
As illustrated in Figure \ref{fig_cam_vis_ir}, the \ref{cam_person_vis} and \ref{cam_person_ir} show visible and infrared images of an identify in the test set.
We use the Grad-CAM \cite{DBLP:conf/iccv/SelvarajuCDVPB17} method to generate attention maps (Figure 6c-f) on them.
Compared with Figure \ref{cam_base_id1_vis}, we can notice that Figure \ref{cam_base_me_id1_vis} captures more id's profile and texture information (such as the pattern on this man's shirt), which are modality-invariant.
For infrared image \ref{cam_person_ir}, it is difficult for human eyes to distinguish the pattern on the shirt, BASE+ME can still perceive the information of this position in Figure \ref{cam_base_me_id1_ir}.
Besides, the consistency of \ref{cam_base_me_id1_vis} and \ref{cam_base_me_id1_ir} is higher than that of \ref{cam_base_id1_vis} and \ref{cam_base_id1_ir}, which demonstrates the auxiliary ability of ME to reduce the modality difference. 


\subsubsection{Analysis of the MAE Loss.}
We show feature distributions of the learned embeddings with t-SNE \cite{van2008visualizing} method in Figure \ref{fig_feature_distribution}. 
Different colors represent different IDs.
The solid square and hollow circle represent two modalities respectively.
From Figure \ref{base} to \ref{base_me_mac} and \ref{base_me_maid}, the MAC loss helps reduce the intra-class differences, and the MAID loss helps expand the inter-class differences.
As shown in Figure \ref{base_me_mac_maid}, when using the MAE loss, embeddings have better distribution with compact intra-class distances and uniformly larger inter-class distances.
Thus, MAE loss helps to generate more effective embeddings for retrieval.

\section{Conclusion}
In this paper, we propose a novel method for the VI-ReID task, the cross-modality transformer (CMTR) network. 
By introducing the modality embeddings (ME), the model can directly perceive characteristics of each modality. 
Furthermore, we design the modality-aware enhancement Loss, which can enhance the ME's learning ability and help to generate better discriminative modality-invariant embeddings. 
The method shows great experimental performance against CNN-based methods on SYSU-MM01 and RegDB datasets. 
We believe that the proposed strategy will provide promising solutions for other cross-modality vision task.


\bibliography{aaai22}
\clearpage

\subsection{Appendix}
\vspace{8pt}

\subsection{A. Comparison on Stride of Patch Generation}
When generating patch sequences, we do this in an overlapping manner, as suggested in the previous research \cite{DBLP:journals/corr/abs-2101-11986, DBLP:journals/corr/abs-2102-04378}.
Specifically, consistent with the input of the base ViT model \cite{DBLP:conf/iclr/DosovitskiyB0WZ21}, the patch size used in our method is $16 \times 16$ ($P=16$), and we conduct comparative experiments with different strides on our cross-modality transformer (CMTR) network.

As shown in Table \ref{tab_stride_comparison}, a series of experiments are implemented with stride $S$ set to $\{16, 14, 12, 10, 8\}$ in turn.
Note that $N$ represents the length of patch sequence.
When $S$ is set to 16 ($S=P$), there is no overlap between patches.
From Table \ref{tab_stride_comparison}, we can notice that this method (Index-1) achieves relatively low performance with 57.38\% Rank-1 and 56.56\% mAP.
As $S$ decreases, the overlapping area between adjacent patches will gradually increase.
In the meantime, the performance of the model is gradually improved.
Compared with the Index-1 model, other models (Index-2,3,4,5) gain +0.95\%,+2.51\%,+3.18\%,+5.20\% Rank-1 and +0.91\%,+2.00\%,+2.90\%,+4.77\% mAP, respectively.
These experimental results verify the effectiveness of the overlapping patch strategy in this VI-ReID task.

With a smaller stride $S$, the model can generate a longer patch sequence with larger $N$ (shown in the second and third columns of the Table \ref{tab_stride_comparison}).
Through overlapping patches and long sequences, the multi-head self-attention modules of the transformer model can learn richer features.
In practice, we use the stride set to 8, so that half of the patches overlap each other, ensuring better performance, while controlling the method’s consumption of computing resources.

\subsection{B. Study of the Parameter $\lambda$ in Objective Function}
In the definition of overall objective function $\mathcal{L}_{overall}$, the weight $\lambda$ is used as a hyperparameter to balance the proportion between losses.
During the experiment, we evaluate the effect of this hyperparameter $\lambda$.

\begin{figure}[t]
\centering
\includegraphics[width=\linewidth]{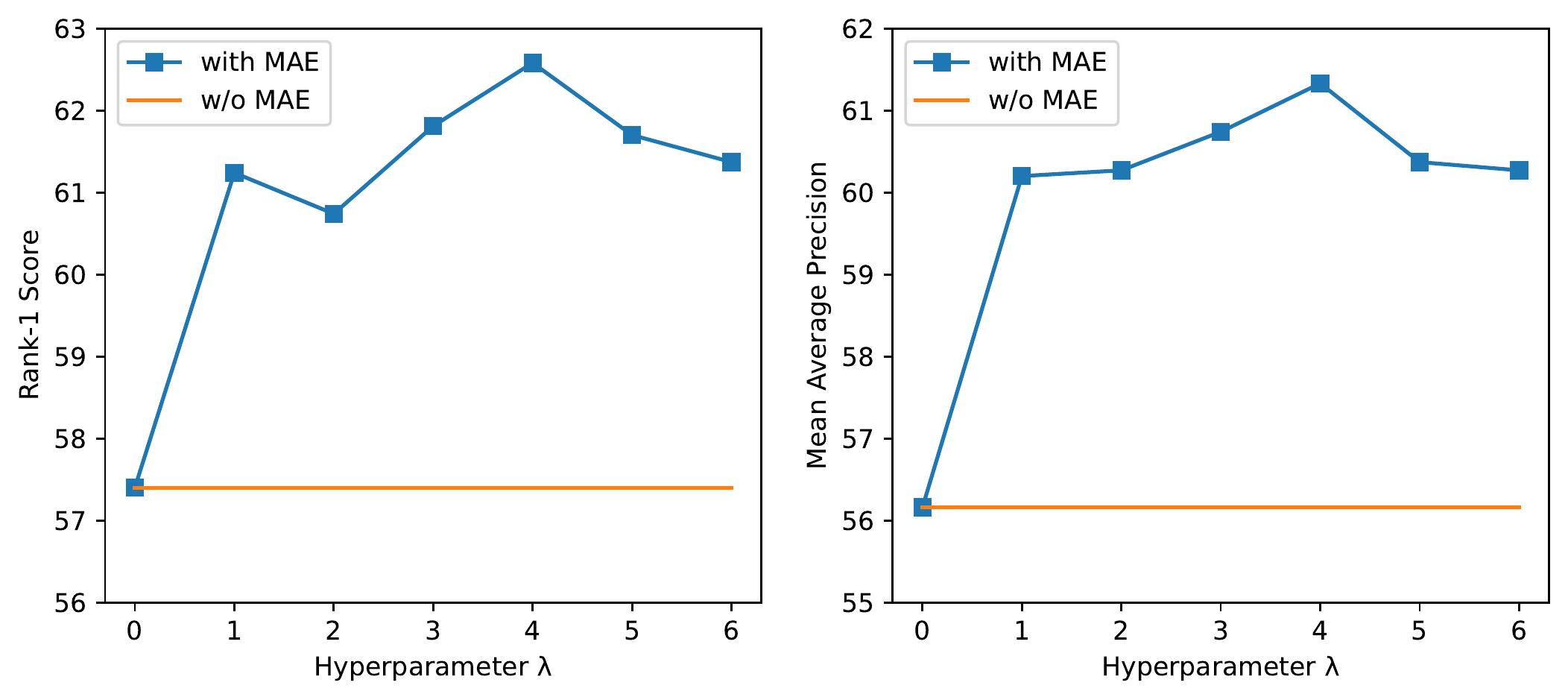}
\caption{Comparison of CMTRs' performance with different settings on hyperparameter $\lambda$ in objective function.}
\label{fig_lambda_study}
\end{figure}

As illustrated in Figure \ref{fig_lambda_study}, the hyperparameter $\lambda$ is set to different values $\{0,1,2,3,4,5,6\}$.
With $\lambda$ set to zero, the network is optimized without (w/o) the proposed modality-aware enhancement loss.
We plot its Rank-1 and mAP line as a benchmark.
By introducing the modality-aware enhancement loss with hyperparameter $\lambda$ greater than zero, we can observe a significant performance improvement on both Rank-1 and mAP, and it is relatively stable with different $\lambda$ values.
The experimental results demonstrate that our method is robust to this weight in objective function.

\subsection{C. Comparative Experiments of Loss Constraints}
There exist some studies that focus on adjusting the distribution of matching embeddings through loss constraints, such as the center loss \cite{DBLP:conf/eccv/WenZL016} and Hetero-Center (HC) loss \cite{DBLP:journals/ijon/ZhuYWZHT20}.
During the experiment, we compare our proposed modality-aware enhancement (MAE) loss with these loss constraints in the cross-modality transformer (CMTR) network to check their performance.

\begin{table}[t]
    \centering
    \fontsize{9}{15}\selectfont
    \setlength{\tabcolsep}{1.7mm}
    \begin{tabular}{c|c|c|c c c c c}
        \hline
        \hline
        Index & Stride & N & R1 & R10 & R20 & mAP & mINP \cr
        \hline
        1 & $S=16$ & 128 & 57.38 & 91.78 & 97.19 & 56.56 & 44.07 \cr
        2 & $S=14$ & 162 & 58.33 & 91.85 & 96.62 & 57.47 & 45.04 \cr
        3 & $S=12$ & 210 & 59.89 & 92.91 & 97.60 & 58.56 & 46.04 \cr
        4 & $S=10$ & 300 & 60.56 & 92.81 & 97.12 & 59.46 & 46.73 \cr
        5 & $S=~~8$ & 465 & \textbf{62.58} & \textbf{93.79} & \textbf{98.01} & \textbf{61.33} & \textbf{49.01} \cr
        \hline
        \hline
    \end{tabular}
    \caption{Experimental results with different strides of patch generation (SYSU-MM01 \& all-search single-shot mode).}
    \label{tab_stride_comparison}
\end{table}

\begin{table}[H]
    \centering
    \fontsize{9}{15}\selectfont
    \setlength{\tabcolsep}{1mm}
    \begin{tabular}{c|l|c c c c c}
        \hline
        \hline
        Index & ~~~~~~~~~~~Methods & R1 & R10 & R20 & mAP & mINP \cr
        \hline
        1 & Baseline+ME & 57.40 & 89.31 & 95.13 & 56.16 & 43.86 \cr
        2 & Baseline+ME+Center & 58.09 & 93.01 & 97.63 & 57.88 & 45.61 \cr
        3 & Baseline+ME+HC & 58.21 & 92.98 & 97.63 & 58.05 & 45.98 \cr
        4 & Baseline+ME+MAE & \textbf{62.58} & \textbf{93.79} & \textbf{98.01} & \textbf{61.33} & \textbf{49.01} \cr
        \hline
        \hline
    \end{tabular}
    \caption{Comparative study on loss constraints under the cross-modality transformer (CMTR) method.}
    \label{tab_loss_constraints_comparison}
\end{table}

As shown in Table \ref{tab_loss_constraints_comparison}, the "Baseline+ME" denotes the baseline ViT network with modality embeddings (ME) optimized by the common $\mathcal{L}_{ID}$ and $\mathcal{L}_{WRT}$ losses.
We add center loss ("Center" for short), Hetero-Center loss ("HC"), and the proposed modality-aware enhancement ("MAE") loss to this basic network (Index-2,3,4 of Table \ref{tab_loss_constraints_comparison}).
From quantitative evaluation results, the center loss and HC loss bring little performance improvement.
The former brings +0.69\% Rank-1 and +1.72\% mAP, the latter brings +0.81\% Rank-1 and +1.89\% mAP.
Compared with "Baseline+ME", the MAE loss shows a prominent effect (Index-4), and its Rank-1 and mAP results are greatly improved.

The traditional loss constraints directly act on extracted embeddings.
However, they do not consider the effective mining and rational utilization of modalities' characteristics and information, which limits their performance.
In contrast, by using modality embeddings (ME), our designed modality-aware enhancement (MAE) loss overcomes the shortcomings of existing losses and provides better guidance for the adjustment of matching embeddings' distribution.

\begin{table*}[t]
    \centering
    \fontsize{9}{14}\selectfont
    \setlength{\tabcolsep}{3.4mm}
    \begin{tabular}{l|c|c c c c|c c c c}
        \hline
        \hline
        \multirow{2}{*}{~~~~~~~~~~~~~~~~Methods} & \multirow{2}{*}{Venue}
        & \multicolumn{4}{c}{\textit{Visible to Thermal}} & \multicolumn{4}{|c}{\textit{Thermal to Visible}} \cr
        & ~ & R1 & R10 & R20 & mAP & R1 & R10 & R20 & mAP \cr
        \hline
        Zero-Padding \cite{DBLP:conf/iccv/WuZYGL17} & ICCV 17 & 17.75 & 34.21 & 44.35 & 18.90 & 16.63 & 34.68 & 44.25 & 17.82 \cr
        SDL \cite{DBLP:journals/tcsv/KansalSWS20} & TCSVT 20 & 26.47 & 51.34 & 61.22 & 23.58 & 25.74 & 50.23 & 59.66 & 22.89 \cr
        BDTR \cite{DBLP:conf/ijcai/YeWLY18} & IJCAI 18 & 33.47 & 58.42 & 67.52 & 31.83 & 32.72 & 57.96 & 68.86 & 31.10 \cr
        MAC \cite{DBLP:conf/mm/YeLL19} & MM 19 & 36.43 & 62.36 & 71.63 & 37.03 & 36.20 & 61.68 & 70.99 & 36.63 \cr
        HSME \cite{DBLP:conf/aaai/HaoWLG19} & AAAI 19 & 41.34 & 65.21 & 75.13 & 38.82 & 40.67 & 65.35 & 75.27 & 37.50 \cr
        D$^2$RL \cite{DBLP:conf/cvpr/WangWZCS19} & CVPR 19 & 43.40 & 66.10 & 76.30 & 44.10 & - & - & - & - \cr
        MSR \cite{DBLP:journals/tip/FengLX20} & TIP 19 & 48.43 & 70.32 & 79.95 & 48.67 & - & - & - & - \cr
        JSIA \cite{DBLP:conf/aaai/WangZYCCLH20} & AAAI 20 & 48.50 & - & - & 49.30 & 48.10 & - & - & 48.90 \cr
        D-HSME \cite{DBLP:conf/aaai/HaoWLG19} & AAAI 19 & 50.85 & 73.36 & 81.66 & 47.00 & 50.15 & 72.40 & 81.07 & 46.16 \cr
        AlignGAN \cite{DBLP:conf/iccv/WangZ0LYH19} & ICCV 19 & 57.90 & - & - & 53.60 & 56.30 & - & - & 53.40 \cr
        FMSP \cite{DBLP:journals/ijcv/WuZGL20} & IJCV 20 & 65.07 & - & - & 64.50 & - & - & - & - \cr
        CMM+CML \cite{DBLP:conf/mm/LingZLRLS20} & MM 20 & - & - & - & - & 59.81 & 80.39 & 88.69 & 60.86 \cr
        X-Modality \cite{DBLP:conf/aaai/LiWHG20} & AAAI 20 & - & - & - & - & 62.21 & 83.13 & 91.72 & 60.18 \cr
        cm-SSFT \cite{DBLP:conf/cvpr/LuWLZLCY20} & CVPR 20 & 65.40  & - & - & 65.60  & 63.80 & - & - & 64.20 \cr
        DDAG \cite{DBLP:conf/eccv/YeSCSL20} & ECCV 20 & 69.34 & 86.19 & 91.49 & 63.46 & 68.06 & 85.15 & 90.31 & 61.80 \cr
        DEF \cite{DBLP:conf/mm/HaoWGLW19} & MM 19 & 70.13 & 86.32 & 91.96 & 69.14 & 67.99 & 85.56 & 91.41 & 66.70 \cr
        Hi-CMD \cite{DBLP:conf/cvpr/ChoiLKKK20} & CVPR 20 & 70.93 & 86.39 & - & 66.04 & - & - & - & - \cr
        HAT \cite{DBLP:journals/tifs/YeSS21} & TIFS 21 & 71.83 & 87.16 & 92.16 & 67.56 & 70.02 & 86.45 & 91.61 & 66.30 \cr
        \hline
        CMTR (Ours) & - & \textbf{80.62} & \textbf{92.93} & \textbf{96.21} & \textbf{74.42} & \textbf{81.06} & \textbf{93.36} & \textbf{96.52} & \textbf{73.75} \cr
        \hline
        \hline
    \end{tabular}
    \caption{Comparison with existing CNN-based cross-modality ReID methods on the RegDB dataset.}
    \label{tab_comparison_regdb_sota_complete}
\end{table*}

\begin{table}[t]
    \centering
    \fontsize{9}{15}\selectfont
    \setlength{\tabcolsep}{1.2mm}
    \begin{tabular}{c|c|c c c c c}
        \hline
        \hline
        \multicolumn{2}{c|}{Study of MAE Loss} & R1 & R10 & R20 & mAP & mINP \cr
        \hline
        \multirow{2}[2]{*}{$\phi_{m}(\cdot)$} 
        & Identity Mapping & 55.59 & 91.92 & 97.49 & 54.83 & 41.61 \cr
        & Fully Connection & \textbf{62.58} & \textbf{93.79} & \textbf{98.01} & \textbf{61.33} & \textbf{49.01} \cr
        \hline
        \multirow{4}[2]{*}{$\mathcal{D}(\cdot,\cdot)$}
        & $\mathcal{L}1$ & 57.65 & 92.50 & 97.43 & 57.10 & 44.48 \cr
        & $\mathcal{L}2$ & 59.40 & 93.66 & 97.88 & 59.09 & 46.69 \cr
        & $Smooth$-$\mathcal{L}1$ & 59.89 & 93.45 & 97.81 & 58.97 & 46.28 \cr
        & $Cos$ & \textbf{62.58} & \textbf{93.79} & \textbf{98.01} & \textbf{61.33} & \textbf{49.01} \cr
        \hline
        \hline
    \end{tabular}
    \caption{Comparison of multiple designs for MAE loss.}
    \label{tab_MAE_design_comparison}
\end{table}

\subsection{D. More Experiments on the Design of MAE Loss}
During the experiment, we compare different implementation schemes of MAE loss, and explore the influence of several designs for the mapping function $\phi_{m}(\cdot)$ and distance measurement method $\mathcal{D}(\cdot,\cdot)$ in Equation \ref{eq_MAC_loss} and \ref{eq_MAID_loss}.

The $\phi_{m}(\cdot)$ is the function to mine the information from modality embeddings (ME).
As shown in Table \ref{tab_MAE_design_comparison}, we compare two implementations, the identity mapping and fully connection.
With the identity mapping, the modality removal operation uses the extracted features to directly subtract the ME of the corresponding modality, which is a hard way to eliminate the learned modalities' characteristics.
Different from this way, the introduction of fully connection layer can further mine helpful information and knowledge in the ME, which can support modality removal in a soft way.
And the experimental results in the Table \ref{tab_MAE_design_comparison} show the positive effect of fully connection. 

As for $\mathcal{D}(\cdot,\cdot)$ in the definition of loss function (Equation \ref{eq_MAC_loss}), we compare various distance calculation methods, including the $\mathcal{L}1$, $\mathcal{L}2$, $Smooth$-$\mathcal{L}1$ and cosine distance.
Our designed MAE loss acts on the features extracted by the model after the BN layer, and the features at this location are constrained by the common ID loss at the same time.
As discussed in paper \cite{DBLP:conf/cvpr/0004GLL019}, the ID loss mainly optimizes the cosine distance.
Thus, our MAE loss should also act as a constraint with distance of the same kind.
The lower part of Table \ref{tab_MAE_design_comparison} shows the experimental results with different distance metrics.
Among all these settings, the cosine distance has a significant advantage, which is also consistent with paper \cite{DBLP:conf/cvpr/0004GLL019}'s analysis and conclusion.

\subsection{E. Complete experimental results on the RegDB dataset with CNN-based methods}
In the text, due to space constraints, the results under all evaluation metrics are not fully displayed in Table \ref{tab_comparison_regdb_sota}.
We show the complete experimental results on the RegDB dataset with existing methods in Table \ref{tab_comparison_regdb_sota_complete}, including the calculated Rank-1,10,20 and mAP.
In addition to Rank-1 (R1) and mAP mentioned in the main text, we can observe that our CMTR method also has notable performance improvement on Rank-10 (R10) and Rank-20 (R20) under these two evaluation modes, which verifies robustness of the method.


\end{document}